\documentclass[sigconf,9pt]{acmart}
\renewcommand\footnotetextcopyrightpermission[1]{}

\usepackage{tikz}
\usepackage{amsmath}
\usepackage{soul}
\usepackage{multirow}
\usepackage{graphicx}
\usepackage{xcolor}
\usepackage[linesnumbered,ruled,vlined]{algorithm2e}
\usepackage{wasysym}
\usepackage{booktabs}
\usepackage{pifont}
\newcommand{\xmark}{\ding{55}}%
\usepackage{hyperref}
\usepackage{listings}
\usepackage{seqsplit}
\usepackage{subcaption}

\usepackage[symbol]{footmisc}

\newcommand\black[1]{\textcolor{black}{#1}}

\newcommand{\pname}{{\texttt{REACT}}\xspace}

\begin{document}

\title{Streaming Video Analytics On The Edge With Asynchronous Cloud Support}

\author{Anurag Ghosh}
\authornote{Work done while at Microsoft Research}
\affiliation{%
   \institution{Carnegie Mellon University}
   \city{Pittsburgh}
   \state{PA}
   \country{USA}}
\email{anuraggh@andrew.cmu.edu}

\author{Srinivasan Iyengar}
\affiliation{%
   \institution{Microsoft Research}
   \city{Bangalore}
   \country{India}}
\email{sriyengar@microsoft.com}

\author{Stephen Lee}
\affiliation{%
   \institution{University of Pittsburgh}
   \city{Pittsburgh}
   \state{PA}
   \country{USA}}
\email{stephen.lee@pitt.edu}

\author{Anuj Rathore}
\authornotemark[1]
\affiliation{%
   \institution{Clutterbot}
   \state{Bangalore}
   \country{India}}
\email{anuj@clutterbot.com}

\author{Venkat N Padmanabhan}
\affiliation{%
   \institution{Microsoft Research}
   \city{Bangalore}
   \country{India}}
\email{padmanab@microsoft.com}



\begin{abstract}



Emerging Internet of Things (IoT) and mobile computing applications are expected to support latency-sensitive deep neural network (DNN) workloads. To realize this vision, the Internet is evolving towards an edge-computing architecture, where computing infrastructure is located closer to the end device to help achieve low latency. However, edge computing may have limited resources compared to cloud environments and thus, cannot run large DNN models that often have high accuracy.


In this work, we develop \texttt{REACT}, a framework that leverages cloud resources to execute large DNN models with higher accuracy to improve the accuracy of models running on edge devices. To do so, we propose a novel edge-cloud fusion algorithm that fuses edge and cloud predictions, achieving low latency and high accuracy. We extensively evaluate our approach and show that our approach can significantly improve the accuracy compared to baseline approaches. We focus specifically on object detection in videos (applicable in many video analytics scenarios) and show that the fused edge-cloud predictions can outperform the accuracy of edge-only and cloud-only scenarios by as much as 50\%. We also show that \texttt{REACT} can achieve good performance across tradeoff points by choosing a wide range of system parameters to satisfy use-case specific constraints, such as limited network bandwidth or GPU cycles.
\end{abstract}

\settopmatter{printfolios=true}
\maketitle
\pagestyle{plain}


\section{Introduction}
Many emerging smart video analytics applications, such as traffic state detection
, health monitoring
, surveillance 
and assistive technology 
require fast processing and real-time response to work effectively. Such applications in built environment monitoring rely on deep learning-based object detection models as a core part of their processing pipeline. These models are compute-intensive 
and tend to have large memory requirements.

Prior works have looked at \textit{offloading} object detection to the cloud~\cite{liu2019edge, chen2015glimpse}. By transferring data, the inference is either entirely or partially offloaded to make use of the compute available in the cloud. However, sending vast quantities of data to the cloud often increases latency, making it unsuitable for near real-time analysis. For intelligent drones~\cite{iyengar2021holistic} or smartphone based driver assistance~\cite{bhandari2018deeplane} to be practical, object detection is needed at low latency without missing any objects. Thus, we believe that improving foundational real-time vision tasks in a manner that is informed by systems considerations would have a beneficial impact on all these applications.

\begin{table}[t]
    \centering
    \small
    \begin{tabular}{l|c|c|c|c} \toprule
         {\bf Features} &\begin{tabular}[c]{@{}l@{}}Our\\Approach\end{tabular} & \begin{tabular}[c]{@{}l@{}}Glimpse\\\cite{chen2015glimpse}\end{tabular} & \begin{tabular}[c]{@{}l@{}}Marlin\\\cite{apicharttrisorn2019frugal}\end{tabular} 
         & \begin{tabular}[c]{@{}l@{}}Edge-Ast.\\ \cite{liu2019edge}\end{tabular}  \\\midrule 
         detection at edge      & \checkmark  & \xmark &  \checkmark  
         &  \xmark \\  \hline
         detection at cloud & \checkmark& \checkmark & \xmark 
          & \checkmark  \\ \hline
         \begin{tabular}[c]{@{}l@{}}n/w variability\\ resilience\end{tabular} & \checkmark & \checkmark & \checkmark 
         & \xmark \\\bottomrule
         
    \end{tabular}
    \caption{A comparison of our approach with existing video analytics techniques.}
    \label{tab:technique}
\vspace{-0.3in}
\end{table}

Edge computing has emerged as an approach to address the latency issue with cloud infrastructure. Small form-factor hardware that are low-cost and consume lower power are often suited for such scenarios. But these often fall short of the heavy computing needs of deep learning models. As such, there has been significant focus on special-purpose devices --- e.g., Nvidia Jetson, Google Coral 
--- optimized to run specific DNN workloads. While edge accelerators provide improved performance over a general-purpose edge computing platform, they are still limited in their support\footnote{Google Coral only supports integer (INT8) operations. Support for some specialized DNN layers/operations is not available in Jetson devices for FLOAT16 and INT8 operations.} compared to cloud-based GPUs. Further, due to system constraints, these approaches run smaller and quantized models at the edge, with lower accuracy, compared to the larger models, with significantly higher accuracy, run on the cloud~\cite{huang2017speed}. 

In this paper, we seek to answer the following research question: 
\textit{Can we have the best of both worlds, i.e., the low latency of the edge models and the high accuracy of the cloud models?} In contrast to cloud-only and edge-only approaches, our key idea is to employ edge-based and cloud-based models in tandem with the cloud resources accessible over a wide-area network that may have high latency. By having \emph{redundant computation} of object detections, we can use cloud-based inferences \emph{asynchronously} to \emph{course correct} edge-based inferences, thereby improving accuracy without sacrificing latency. Table~\ref{tab:technique} distinguishes our work from the prior work involving cloud-only and edge-only approaches.

We exploit this arbitrage between cloud and edge as the performance disparity would remain for years ahead. Past works~\cite{huang2017speed, zhu2018visdrone} in the computer vision community have proposed using model ensemble approaches. However, they combine detections from different models with comparable performance and do so on the same frame without latency considerations. \pname{}'s novel \emph{fusion algorithm} in contrast combines higher accuracy cloud-based detections on recent frames with current inference on the less-accurate edge detections while removing irrelevant stale results from the cloud.

\begin{figure}[t]
    \centering
    \includegraphics[width=2.4in]{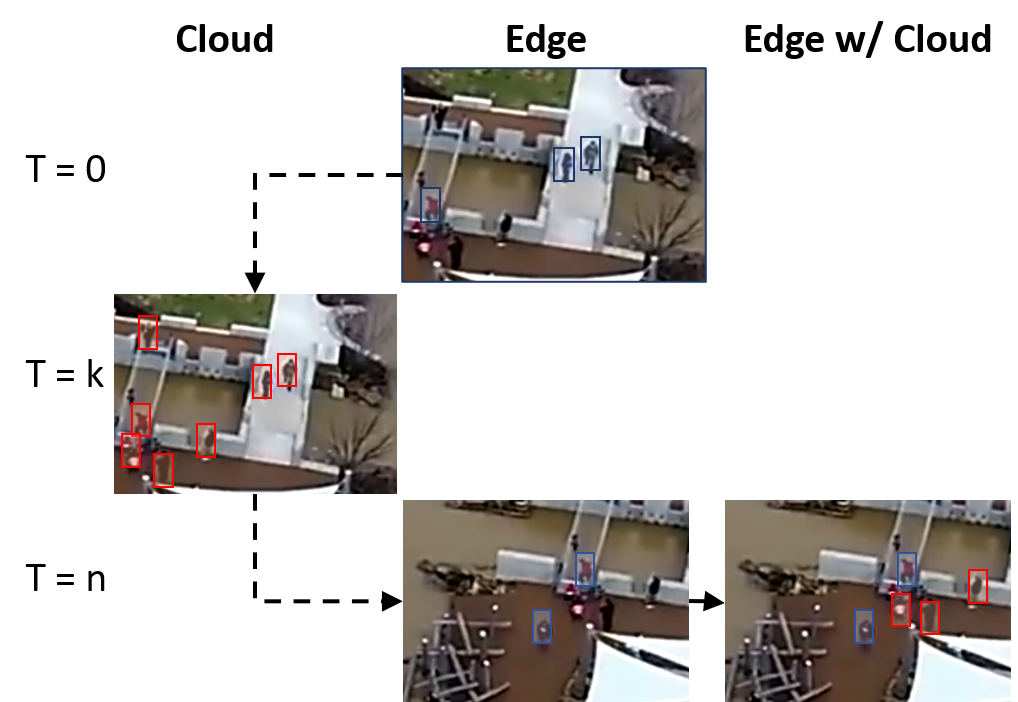}
    \caption{Illustrates the efficacy of asynchronous cloud response to improve edge performance. Note that objects are undetected on edge but detected in the cloud. Thus, cloud responses can be cascaded to improve system performance.} 
    \label{fig:motivating_example}
\vspace{-0.25in}
\end{figure}

Figure~\ref{fig:motivating_example} illustrates how redundant computation helps improve overall accuracy for object detection. The models detect people on a flood-affected riverbank area collected from an intelligent drone at two different points in time. As shown, a cloud-based detection model achieves higher accuracy but comes with significant latency, wherein the results of a frame sent at $t=0$ are obtained at $t = k$. On the other hand, the edge-based detection model has \textit{lower accuracy}, as several humans are not detected. Note that at $t=n$, even though the scene has changed, some people are still common across the current and previous frames. However, the edge model still does not detect these people. Moreover, edge results may be false positives. Thus, we use cloud-based models to improve the overall accuracy by considering detections from the accurate cloud model at time $k < n$ and \textit{merging} these with the frame at $t=n$ on the edge. We note that this merge operation is not trivial. We need to consider cases where both detectors don't agree with each other. Moreover, combining results will not work if the edge receives a cloud response after all the objects of interest within the frame change. It is necessary to ensure that approaches must work in highly dynamic environments, where objects of interest change frequently.

In this paper, we describe \pname{} --- our system that builds on these intuitions to exploit cloud's accuracy with the low latency of the edge. Below, are our contributions.

\noindent
    {\bf REACT System Design:} 
    We designed an edge-cloud video pipeline system capable of exploiting the performance gap of object detection models between the cloud and the edge. Our approach is designed to scale to multiple edge devices and is resilient to network variability. Finally, we develop APIs that edge-based systems can use to leverage cloud-based models and improve overall accuracy.
    
    \noindent
    {\bf Edge-Cloud Fusion Algorithm:} We develop a novel fusion algorithm that combines predictions from edge and cloud object detection models to achieve higher accuracy than edge-only and cloud-only scenarios. To the best of our knowledge, we are the first to leverage redundant computations to improve the accuracy of on-edge object detection.   
    
    \noindent
    {\bf Real-world Evaluation:} We evaluate \pname on two challenging real-world datasets --- data collected from car dashcams~\cite{che2019d} and drones~\cite{zhu2020vision}. These datasets span different cities and exhibit high variations in scene characteristics and dynamics. Our results show \pname can significantly improve accuracy by 50\% over baseline methods. Further, \pname can tradeoff edge and cloud computation while maintaining the same level of accuracy. For instance, by reducing the edge detection frequency by a fourth (from every 5th frame to every 20th frame) and increasing cloud frequency (from every 100th frame to 30th frame), \pname can achieve similar accuracy. 

    \noindent
    {\bf Scalability and Resilience Analysis:} We analyze the scalability of our approach and show \pname can support 60+ concurrent edge devices on a single machine with a server-class GPU.
    We also show that \pname is resilient to network variability. That is, it can function on varying network conditions and leverages cloud models when feasible.  
    We evaluate \pname over different network types (WiFi and LTE) with varying latency using a network emulator. Our results show that even with varying response latency from the cloud, \pname performs better 
    than the edge-only scenario, 

\section{Background}
\label{sec:background}

In this section, we provide background on video-based applications and challenges in cloud or edge-based video analytics applications.

Video-analytics systems collect rich visual information that offers insights into the environment. These systems can be broadly categorized as: (i) devices that send all video to the cloud for processing, and (ii) devices that have limited processing capabilities constrained by its small form-factor, cost, or energy. In this case, the video processing can be split between the device and the cloud. That is, the device can perform either some or possibly all the processing before it sends the video to the cloud. Deep learning inference for object detection forms the core aspect of such systems.

Since deep learning is compute-intensive, existing systems typically send data to the cloud for processing. However, cloud analysis may incur significant delays and may be unsuitable for live applications. \emph{Edge computing} has emerged as an alternative to complement the cloud, where data processing is done close to the devices to avoid these delays. A variety of edge computing architectures exist, depending on where the edge servers are located relative to the end-devices~\cite{satyanarayanan2009case}. Our work assumes the edge device is of low latency, and limited computing capabilities, such as hubs in smart homes, routers, and mobile phones and IoT devices such as intelligent drones and wearable VR headsets. We assume that some form of resource constrained AI-based workloads can be run on these edge devices. Modern devices like Raspberry Pi or Jetson are devices are capable of running lightweight models~\cite{sandler2018mobilenetv2} with a smaller memory footprint. Pairing specialized accelerators (such as Google Coral or Intel Movidius) speeds up the inference time of small models without affecting accuracy for a class of model. Unfortunately, larger deep learning models (having higher accuracy than smaller models) are still not within the latency and memory budget of these devices. Larger models require cloud GPU resources, but this comes at the cost of network delays. This is unacceptable for live and streaming applications. In summary, edge processing provides a latency advantage but there remains a significant accuracy gap between real-time prediction on an edge device and offline prediction in a resource-rich setting~\cite{li2020towards}. Our goal in \pname is to leverage cloud processing in tandem with edge processing to bridge the accuracy gap while preserving the latency advantage of edge processing.

\section{REACT Design}
\label{sec:systemdesign}





\begin{figure*}[t]
\begin{tabular}{cc}
\includegraphics[width=3in]{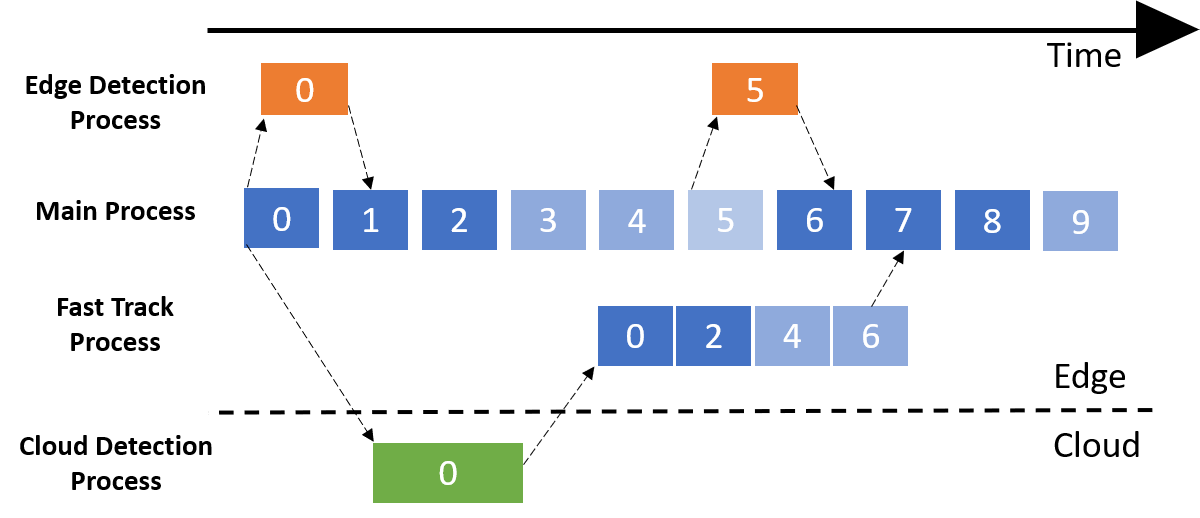} &      \includegraphics[width=4in]{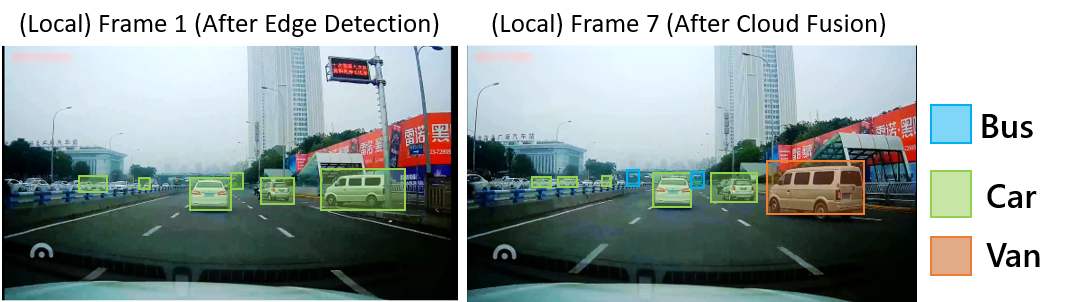}  \vspace{-0.2in}\\
(a) & (b)
\end{tabular}
\caption{(a) \pname{} System Process Flow. Orange and Green boxes indicate edge and cloud detections, respectively. Tracking performance degrades with streaming age, which is indicated by the lighter shades of the color blue. It should be noted that Cloud Detection and Fast Track are both asynchronous events. (b) \pname{} uses \textit{asynchronous} cloud detections to correct the box labels and detect more objects.}
\label{fig:pipeline}
\end{figure*}

 For real-time edge inference, we propose a system that uses an edge-cloud architecture while retaining the low latency of edge devices but achieving higher accuracy \black{than an edge-only approach}. 
In this section, we discuss how we leverage the cloud models to influence and improve edge results. 

\noindent \textbf{Basic Approach:} It is known that
video frames are spatiotemporally correlated. Typically, it is sufficient to invoke edge object detection once every few frames. As illustrated in Figure~\ref{fig:pipeline}(a), edge detection runs every 5th frame. As shown in the Figure, to interpolate the intermediate frames, a comparatively lightweight operation of object tracking can be employed. 
Additionally, to improve the accuracy of  inference, select frames are \emph{asynchronously} transmitted to the cloud for inference. Depending on network conditions (RTT, bandwidth, etc.) and the cloud server configuration (GPU type, memory, etc.), cloud detections are available to the edge device only after a few frames. The newer cloud detections, which were previously undetected, can be brought to the current frame using another instance of an object tracker running on the past buffered images. Video frames retain the spatial and temporal context depending on scene and camera dynamics. Our key insight is that these asynchronous detections from the cloud can help improve overall system performance as the scene usually does not change abruptly. See Figure~\ref{fig:pipeline}(b) for a visual result of the approach.

\noindent \textbf{Challenges:} Nevertheless, designing a system that utilizes the above approach would require addressing several challenges. First, combining the detections from two sources, i.e., local edge detections and the delayed cloud detections is not straightforward. Each of these two detections contain separate list of objects represented by a $\langle$\textit{class\_label}, \textit{bounding\_box}, \textit{confidence\_score}$\rangle$ tuple. A fusion algorithm must consider several cases -- such as class label mismatch, misaligned bounding boxes, etc. -- to consolidate 
\black{the edge and cloud detections into} a single list. Second, 
\black{some or all of} the cloud objects may be ``stale'', \black{outside the current edge frame. The longer it takes to perform fusion, the greater the risk of such staleness, especially}  if the scene changes rapidly. Thus, \black{to minimize this risk,} once the old cloud annotations are received, they must be quickly processed at the edge to help with the current frame.

Another challenge when running detection models on live videos at the edge is minimizing resource utilization while maintaining detection accuracy. Previous studies with edge-only detection systems have shown that running a deep neural network (DNN) for every frame in a video can drain system resources (e.g., battery) quickly~\cite{apicharttrisorn2019frugal}. In our case, with a distributed edge-cloud architecture, several resource constraints need to be simultaneously considered. For example, cloud detections are more accurate as one can run computationally expensive models with access to server-class GPU resources. However, bandwidth constraints or a limited cloud budget might restrict their use to once every few frames. Moreover, if the scene change is insignificant, it would be prudent not to invoke object detections at the edge and the cloud.  On the contrary, for more dynamic scenes, increasing the frequency of edge detection might result in excessive heat generation from the modest GPUs used on edge devices leading to throttling. 

Next, we present our system called \texttt{REACT}, which overcomes the above challenges. Primarily, \texttt{REACT} consists of three components -- i) \texttt{REACT} Edge Manager, ii) Cloud-Edge Fusion Unit, iii) \texttt{REACT} Model Server. Below, we describe them in more detail.

\subsection{\texttt{REACT} Edge Manager}

The \texttt{REACT} Edge Manager (REM) consists of different modules, and put together, enables fast and accurate object detection at the edge.



\noindent \textbf{Change detector:} 
Previous studies have shown that running a object detection on every frame in a video can drain system resources (e.g., battery) quickly~\cite{apicharttrisorn2019frugal}. REM provides two parameters,  i.e., the detection frequency at the edge ($k$) and the cloud ($m$) -- to modulate the number of frames between object detection. Intuitively, if there is little object displacement across frames, running detection models frequently will lead to wastage of resources. REM employs a change detector that computes the \textit{optical flow} 
on successive frames. This represents the relative motion of the scene consisting of objects and the camera, similar to~\cite{apicharttrisorn2019frugal, chen2015glimpse, jiang2018chameleon}. Thus, the object detection invocations will only occur at a detection frequency of every $k^{th}$ and $m^{th}$ frame at the edge and the cloud, respectively, if this motion is greater than a pre-decided threshold. 

\noindent \textbf{Edge Object Detector:}  
Every $k^{th}$ frame, REM triggers the edge object detector module, which in turn outputs a list of  $\langle l, p, c \rangle$ tuples. Here, $l$ and $c$ are class labels (e.g., cars, person) and confidence scores (between 0 and 1) associated with the detected objects, respectively. 
$p=(x, y, w, h)$ represents the bounding box  for each of the detected objects, where  $x, y$ is the center coordinate of the object; $w, h$ is the width and height of the bounding box. 
To avoid multiple bounding boxes for the same object, we use Non-max suppression, which removes locally repeated detections.



\noindent \textbf{Main Object tracker:}
REM employs an CPU-based object tracker, a computationally cheaper technique, between frames for which the object detections are available. For example, a CSRT~\cite{lukezic2017discriminative} tracker can process images at $>$40 fps (on Nvidia Jetson Xavier).
However, as the quantum of associated displacement of objects increases, the tracker accuracy also reduces. The tracker module accounts for this degradation by multiplying every tracked object's confidence scores by a decay rate $\delta \in [0,1]$. As the confidence scores reduce with every passing frame with this multiplier, the module sweeps over the list of objects to discard the ones with lower confidence scores (i.e., $c<0.5$).

\noindent \textbf{Cloud communicator:} 
The REM consists of a communication module responsible for sending every $m^{th}$ frame (cloud detection frequency) to the cloud and receive the associated output annotations. Similar to edge detections, the cloud annotations consist of a list of $\langle l, p, c \rangle$ tuples. Since the cloud can execute larger object detection models, it provides better accuracy over lightweight models running at the edge. The communication module transmits frames asynchronously to the cloud. Again, the cloud detection frequency is based on objects' motion and leverages the change detector module. If the change is below threshold, we do not transmit frames to the cloud for object detection. 
As the cloud always processes an older frame due to network latency, the predictions might become stale \black{(i.e., fall outside the frame)} by the time it reaches the edge.  

\subsection{REACT Model Server}
The \texttt{REACT} Model Server's primary goal is to respond to edge inference requests by executing the object detection models \black{on the cloud} and sending annotations of the detected objects \black{back to the edge device}. The server may be shared across numerous edge devices to handle multiple requests at any given time. A request queue is maintained with multiple worker threads (parameterized by $num\_workers$) to maximize throughput while \black{adhering to a} latency constraint. Server class GPU architectures can efficiently operate in parallel on a batch of images (say, $batch\_size$ image tensors) that are dispatched together for inference. Requests are preprocessed and batched by the worker threads,
and a batch is sent for inference to the GPU(s) either when a batch has $batch\_size$ images for inference or when a $max\_delay$ wait threshold is reached. Optimal parameter choices depend on the GPU hardware characteristics and the distribution of incoming requests. For simplicity, we do not consider dynamic batching scenarios.

\begin{algorithm}
\small
\SetAlgoLined
\DontPrintSemicolon
\SetKwFunction{LinearSumAssignment}{LinearSumAssignment}
\SetKwFunction{RemoveOldDetections}{
RemoveOldDetections}
\SetKwFunction{GetDetectionSource}{GetDetectionSource}
\SetKwFunction{ComputeIOU}{ComputeIOU}
 $M$=[][]\;
 $det\_source = \GetDetectionSource(objects_{new})$\; 
 $objects_{current} = \RemoveOldDetections(objects_{current},det\_source)$\;
 \For{ $o_c \in objects_{current}$} 
 {
     \For{$o_n \in objects_{new}$} 
    {
      $iou=\ComputeIOU(o_c.bbox, o_n.bbox)$\;
    
      \eIf{$iou>=threshold$}{
            $M[o_c][o_n]=iou$\;
        }{
            $M[o_c][o_n]= 0$\; 
        }
    }
 }
 $curr\_objs, new\_objs = \LinearSumAssignment{M}$\;
 $updated\_curr\_objs = []$\;
\For{$o_c, o_n \in zip(curr\_objs, new\_objs)$}{
\eIf{$M[o_c][o_n] != 0$}{
    $o = \{\}$\;
    \If{$det\_source == ``cloud"$}{
        $o.label = o_n.label$\;
        $o.bbox = o_c.bbox$\; 
    }
    \If{$det\_source == ``edge"$}{
        $o.label = o_c.label$\;
        $o.bbox = o_n.bbox$\; 
    }
    $o.score = o_n.score$\;
    $o.score = decay(o.score)\;$
    $o.last\_det\_source = det\_source$\;
    $updated\_curr\_objs += o$\;
}
{
$o_n.last\_det\_source = det\_source$\;
$updated\_curr\_objs += o_n$\;
}
}
\Return $updated\_curr\_objs$\;
\caption{Edge-Cloud Fusion Algorithm}
\label{algo:fusion}
\end{algorithm}

\subsection{Edge-Cloud Fusion Unit}
\label{sec:fusion}


The detections from the edge are available for immediate use. However, the detections received from the cloud are delayed and do not belong to the current frame. To use these detections to improve the current frame's detection, we \emph{fast track} cloud object predictions. Here, we start a new instance of the tracker on a new process separate from the main tracker. Specifically, we initialize this instance with cloud predictions and track the objects on every alternate frame, until it is current. This stride can be increased at the cost of decreased localization accuracy, in practice, we saw tracking on every alternate frame had good accuracy and speed.

The output of DNN models from the edge and cloud is different. Both may detect the same object, the bounding boxes, but the confidence scores, and sometimes the labels may differ due to the model quirks. It is also possible that either the edge or cloud model may fail to detect some objects. Our goal is to combine the edge and cloud predictions to avoid repeated instances of the same object, while adding previously undetected ones.

We develop a novel bounding box fusion algorithm to combine cloud-edge predictions. Many box fusion/selection techniques, such as say non-maximum weighted (NMW)~\cite{zhou2017cad} or NMS, combine predictions based on class labels and considers a match if the overlap of the bounding box is high for the same class. If the labels are different, these techniques will consider it as two different objects. Our analysis showed edge detection models were able to localize objects correctly but often had false positives, i.e., assigned class labels incorrectly. Using above techniques would cause the same object to be considered twice. 

Our box fusion technique works as follows. In the edge, we maintain a current list of objects (in the form of tuples described earlier) for the present frame. Whenever any new detections, either from the cloud or the edge, are available, we first delete the old objects from the current list that were last submitted by the same detection source.  For example, we delete old objects detected by the cloud (or edge) when newer cloud (or edge) detections are available. 

Next, we create an Intersection over Union (IoU) matrix that indicates the overlap between current objects and the detections received. IoU is the ratio of overlapped area with the union of the area between the two sets of objects. 
Any value smaller than a threshold ($\ge 0.5$) is set to 0.
We then perform a linear sum assignment~\cite{burkard1980linear}, which matches two objects with the maximum overlap. This matrix provides a list of objects that were already present in the current object list. We modify the confidence values, bounding box, and class label based on the new detections' source. For example, objects from the cloud obtained from running bigger models will be more accurate in predicting the class correctly. We present the pseudo-code to determine the merging of the boxes  in Algorithm~\ref{algo:fusion}. 

\section{Implementation}

\noindent \textbf{\texttt{REACT} Edge Manager:} 
Our implementation uses OpenCV to receive a stream of video images. Further, the tracking module is built upon OpenCV's object tracker API.  We train two object detection models (MobileNetV2-SSD~\cite{sandler2018mobilenetv2} and TinyYOLO~\cite{redmon2016you} ) for the edge scenario. We also deployed these models on an Nvidia Jetson Xavier device for inference. 

\noindent \textbf{\texttt{REACT} Model Server:} Our object detection models (Faster R-CNN~\cite{ren2015faster} and RetinaNet~\cite{lin2017focal}) are trained using the mmdetection~\cite{mmdetection} library, and for training CenterNet~\cite{zhou2019objects} models we utilize the official implementation written in Pytorch. Pytorch's default \texttt{\seqsplit{object\_detector}} handler only supports torchvision models, so we implemented custom handlers for generic mmdetection models and the CenterNet model to serve them on the cloud server using Torch Serve. We created two handlers as the mmdetection library and the CenterNet library expose and utilize very different model initialization, preprocessing and postprocessing programming paradigms in their implementations. 
Thus, \pname can be used with newer object detection algorithms in the future by modifying the sample handlers for reflecting model specific changes. We serve these models as HTTPS/JSON endpoints over an API.



\noindent \textbf{\pname{} fusion API:}
We expose two classes \texttt{\seqsplit{CloudServerInference}} and \texttt{\seqsplit{ReactEdgeInference}}. \texttt{\seqsplit{CloudServerInference}} class can be instantiated by providing the address of the HTTP endpoint, the image resolution, and the model along with optional frequency parameter, tracker type and number of tracker threads (if used in Server only inference mode). To instantiate \texttt{\seqsplit{ReactEdgeInference}} we specify parameters such as the model to run, image resolution, tracker type, number of tracker threads, the frequency parameters, and \texttt{\seqsplit{CloudServerInference}} object to use. Both the classes expose a \texttt{get\_annotations} method and use the image as input, returning the annotation output as a JSON object. We believe this API design facilitates adoption by application developers due to it's simplicity and ease of use.







\section{Evaluation Methodology}

In this section, we give a detailed description of the datasets used and the evaluation setup.

\subsection{Dataset Description}
\label{subsec:dataset}
We extensively evaluate the proposed system's efficacy on two datasets in built environment monitoring domain highlighting its potential in different use cases (drone-based surveillance and dashcam-based driver assist). Both these datasets are popular and are among the largest available dataset for edge-based object detection.
These datasets are quite challenging as they exhibit significant scene change and have a varied number and size of objects. 
Table~\ref{tab:dataset_summary} provides a summary of the two datasets.  


\noindent \textbf{$D^{2}$-City~\cite{che2019d}:} 
The video dataset is created from front-facing car dashcams and captures the dynamic complexity of real-world traffic conditions. The dataset is crowdsourced from passenger vehicles registered on DiDi's platform and intended for improving vision technologies, driving intelligence, and similar use cases. It has 1000 driving videos taken in five different cities under various scene conditions and video resolution. Objects in each video frame are annotated and include their bounding boxes and class ids. 


\noindent \textbf{VisDrone~\cite{zhu2020vision}:}
The videos in the dataset are captured using drones flown over different cities under various weather and lighting conditions. It contains 79 video clips with around 1.5 million manually annotated objects. We use the evaluation protocol followed in  VisDrone-VDT 2018 challenge for video object detection task~\cite{zhu2018visdrone}, which focuses on detecting specific objects (e.g., pedestrian, car, van) taken from drones. The object detection task in these videos is considered challenging due to the density of really small objects, dynamic scene conditions, and drones' movement. 


\begin{table}[t]
\small
\centering
\begin{tabular}{c|c|c|c|c}
\toprule
\textbf{Name} & \textbf{Type} & \textbf{\begin{tabular}[c]{@{}c@{}}Size (\#videos, \\ \#frames)\end{tabular}} & \textbf{\begin{tabular}[c]{@{}c@{}}\# of\\ Classes\end{tabular}} & \textbf{Remarks} \\ \midrule
VisDrone & Drone & \begin{tabular}[c]{@{}c@{}}(79, \\ 33.3K)\end{tabular} & 12 & \begin{tabular}[c]{@{}c@{}}Altitude,\\ View Angle\end{tabular} \\
\hline
$D^{2}$-City & \begin{tabular}[c]{@{}c@{}}Dash \\ Cam\end{tabular} & \begin{tabular}[c]{@{}c@{}}(1000, \\ 700K)\end{tabular} & 8 & \begin{tabular}[c]{@{}c@{}}Varied object\\ sizes\end{tabular} \\ \bottomrule
\end{tabular}
\caption{Summary of the datasets.}
\vspace{-0.3in}
\label{tab:dataset_summary}
\end{table}

\subsection{Performance Metrics} 
\label{perfmetrics}

We use \textit{mean average precision at intersection over union (IoU) = 0.5} 
($mAP@0.5$)
--- a popular metric used for object detection tasks (See Pascal VOC challenge~\cite{everingham2010pascal}).  Note that the IoU measures the ratio of the intersection area and the area of union of the predicted bounding box and ground truth bounding box. 
Thus, a prediction is considered a true positive if the predicted label matches the ground truth, and the IoU is greater than or equal to the threshold ($\ge$ 0.5). 

\subsection{Evaluation Setup}
In this section, we discuss the training process, baseline techniques and environment.

\subsubsection{Model selection and training}
We use a combination of deep learning models to evaluate our approach,  where we execute different models on the edge and cloud.  For our cloud-based models, we use Faster-RCNN~\cite{ren2015faster}, RetinaNet~\cite{lin2017focal}. For edge models, we use TinyYOLO~\cite{redmon2016you} and MobileNetV2-SSD~\cite{sandler2018mobilenetv2}.
Table~\ref{tab:model_summary} provides a summary of the different models. To train our models, we follow the protocols described in the $D^{2}$-City and VisDrone datasets. 
As these datasets are released as part of ongoing challenges, the test set annotations are not publicly available. Hence, we evaluate our models on the released validation data set. For our validation dataset during training, we use 15\% from the train data set to tune the hyper-parameters and select the final model.  


\subsubsection{Baseline Techniques}
We use the following baseline techniques to compare with our proposed approach.

\noindent \textbf{Edge-only Inference:} Here, we run the object detection only at the edge and do not offload detection tasks to cloud resources. Prior systems such as \textit{Marlin}~\cite{apicharttrisorn2019frugal} employ a similar strategy (see Table~\ref{tab:technique}) along with domain specific improvements (in AR/MR) to reduce energy costs. We use lightweight detection models --- TinyYOLO and MobilNetV2-SSD --- as they consume less memory and computation and are well suited for resource-constrained edge devices. 
From hereon, we refer to this baseline as \textit{edge-only}.

\noindent \textbf{Cloud-only Inference:} For this baseline, we run the object detection task on the cloud. The edge is a thin client that offloads the detection tasks to the cloud while using a tracker to compensate for intermediate frames. The performance of this baseline setup is comparable to existing systems
such as \textit{Edge-Assisted}~\cite{liu2019edge} and \textit{Glimpse}~\cite{chen2015glimpse} (See Table~\ref{tab:technique}) that offloads trigger frames to the cloud and uses an optical flow based object tracking method to update the object bounding boxes on the other frames. Note that cloud-only inference suffers from higher network delays compared to the edge-only scenario~\cite{liu2019edge}. Such high network latency may be undesirable for latency-sensitive applications as dynamic changes in scenes may render responses from the cloud unusable. We use computationally expensive detection models on cloud, namely RetineNet, Faster RCNN, and CenterNet, due to their good performance. From hereon, we refer to this baseline as \textit{cloud-only}.
    

\noindent \textbf{Every Frame Edge Inference:} In this scenario, we compare \texttt{REACT} with the case where one can run detectors using edge models (TinyYOLO and SSD-MobileNetv2) on every frame. Unlike the edge-only baseline, we do not  interpolate predictions with any tracker. In practice, this baseline is infeasible as edge devices cannot run detections on all frames due to latency and energy constraints. We call these baselines \textit{ef-edge-det (tinyyolo)} and \textit{ef-edge-det (ssdmv2)}. We \emph{do not} compare with Every Frame Cloud Inference as it neither meets the computational budget nor the latency budget.

\begin{table}[]
\small
\centering
\begin{tabular}{c|c|c|c}
\toprule
\textbf{Detector} & \textbf{Backbone} & \textbf{Where} & \textbf{\#params} \\ \midrule
Faster R-CNN & ResNet50-FPN & Cloud & 41.5M \\ RetinaNet & ResNet50-FPN & Cloud & 36.1M \\ 
CenterNet & DLA34 & Cloud & 20.1M \\ 
TinyYOLOv3 & DN19 & Edge & 8.7M \\
SSD & MobileNetV2 & Edge & 3.4M \\ \bottomrule
\end{tabular}
\caption{Summary of models used.}
\label{tab:model_summary}
\vspace{-0.3in}
\end{table}

\subsubsection{Network Emulation}
We use \textit{Mahimahi}~\cite{netravali2015mahimahi} and \textit{traffic control (tc)} Linux utility to emulate different network traffic, in particular, LTE and WiFi. For LTE, we use the Verizon LTE uplink and downlink traces in MahiMahi to emulate LTE link between the edge and cloud~\cite{netravali2015mahimahi} (hereon, we refer it as \textit{LTE}).  
For WiFi, we throttle the traffic to 24Mbps and also introduce delay of 30ms and 50ms using \textit{tc}. Hereon, we refer them as \textit{WiFi (30 ms)} and \textit{WiFi (50 ms)}. Thus, we emulate three different network conditions between the client and the server. Unless stated otherwise, we report our results using the \textit{WiFi (30 ms)} network. 


\section{Experimental Results}



In this section, we compare \pname{} with other baseline techniques. We also study the impact of network conditions and the tradeoff opportunities from adjusting the detection frequency at both the cloud and the edge. Further, we evaluate the scalability of our approach and its performance on an edge accelerator device. 


\begin{figure*}[!htb]
\minipage{0.62\textwidth}
    \begin{tabular}{cc}
    \includegraphics[width=0.45\linewidth]{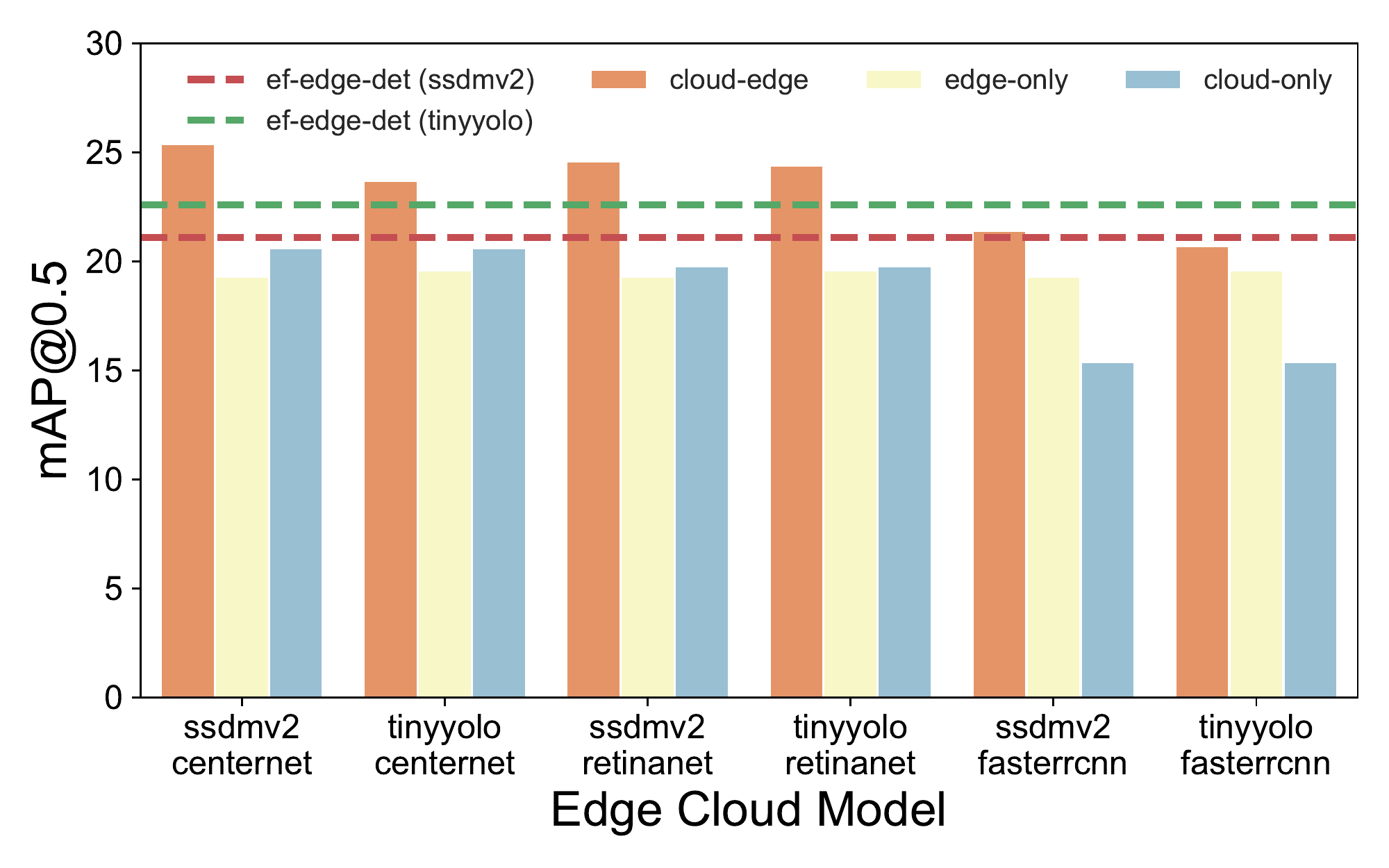} &
    \includegraphics[width=0.45\linewidth]{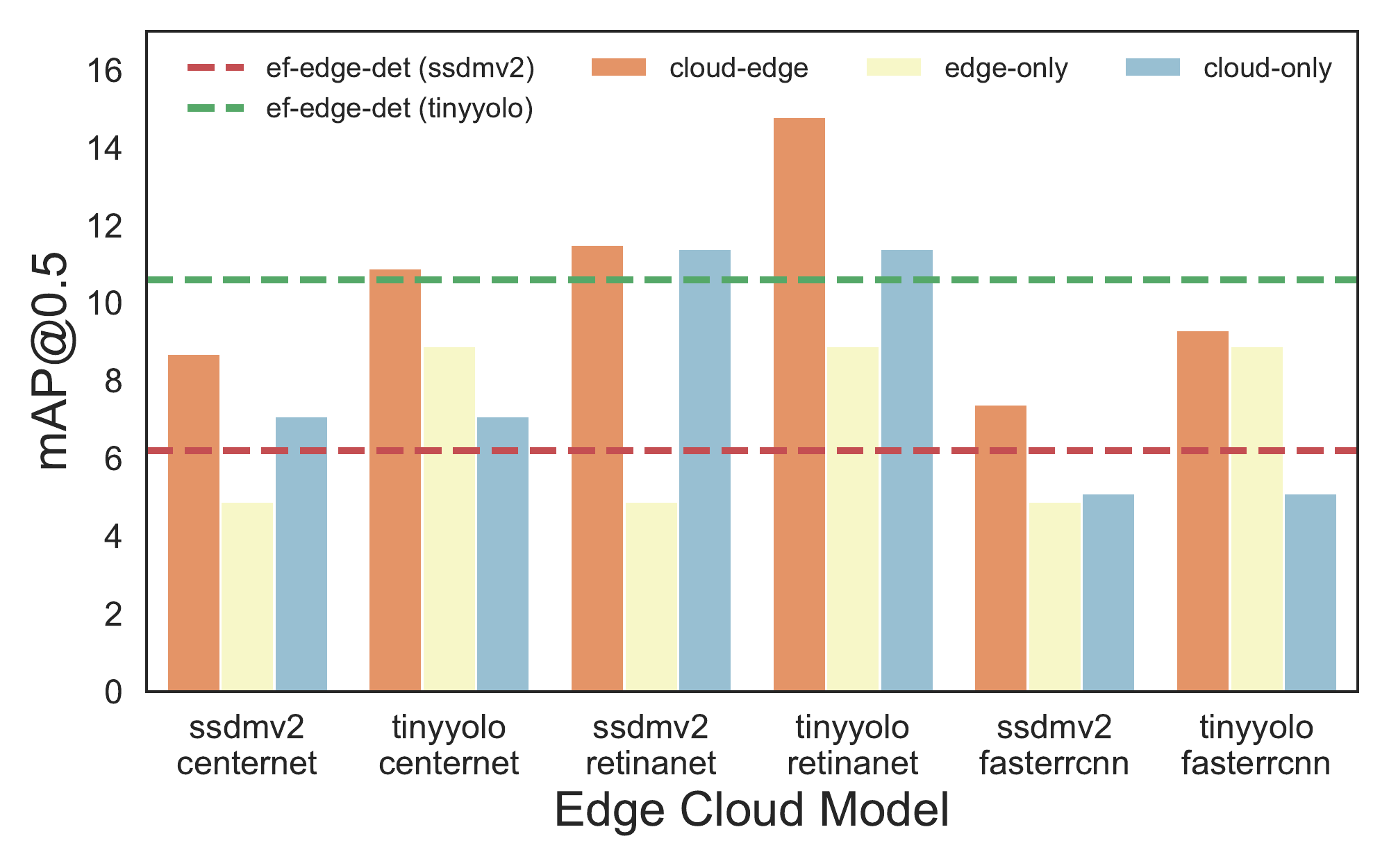}\\
    (a) D2-City & (b) VisDrone\\
    \end{tabular}
    \caption{Baseline comparison of \pname{} across the two datasets.}
    \label{fig:cloudpro}
\endminipage\hfill
\minipage{0.35\textwidth}
    \centering
    \includegraphics[width=0.80\linewidth]{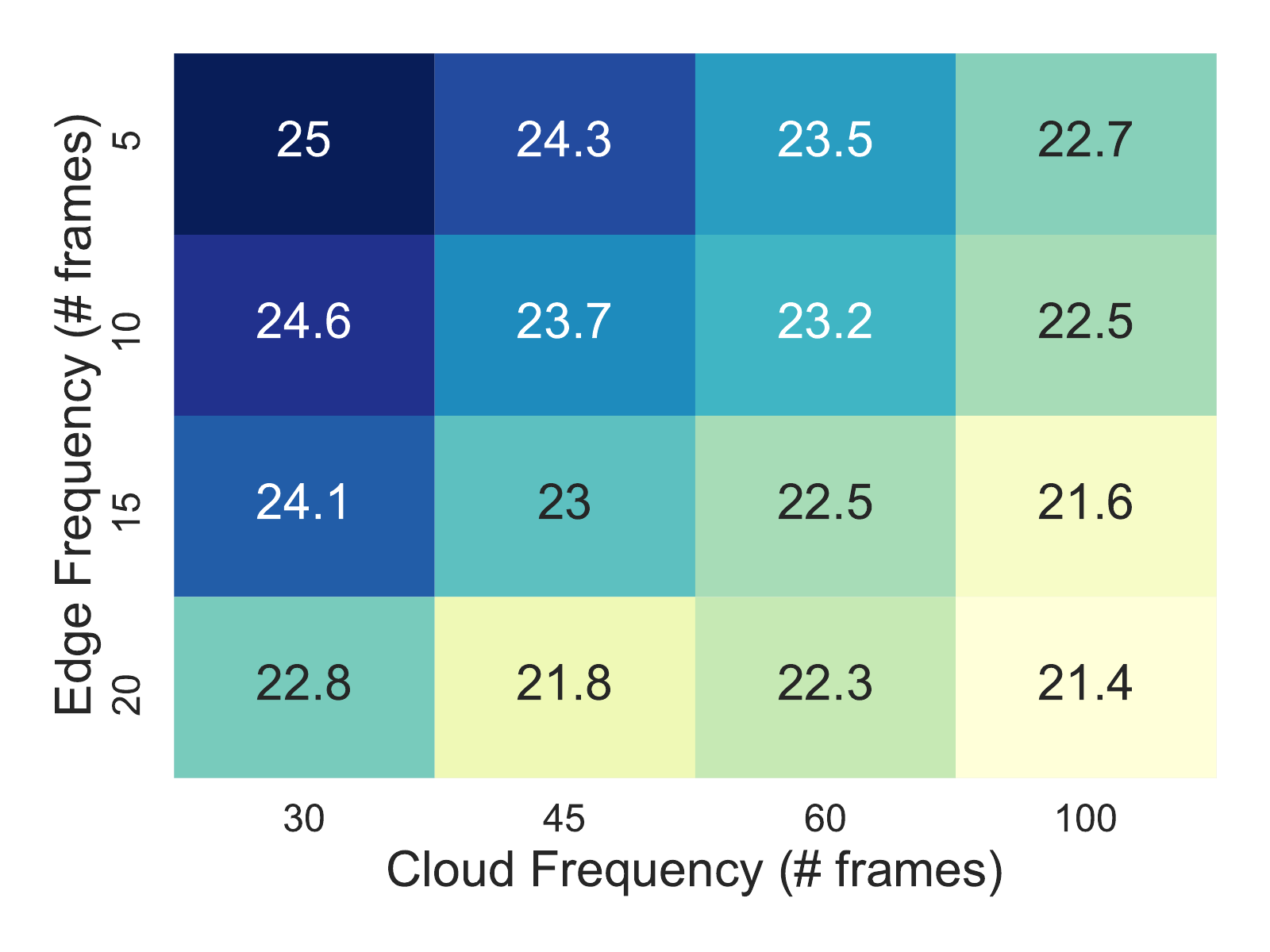}
    \caption{Detection frequency tradeoff.}
    \label{fig:heat}
\endminipage
\end{figure*}

\subsection{Performance Comparison}
\label{subsec:baseline-comp}

We first evaluate how \pname{}'s  use of redundant detections running asynchronously on the cloud help achieve low latency and improves accuracy. 
In this experiment, we set the edge and cloud object detection frequency to 5 and 30, respectively. 
We compare \pname{} to our three baseline approaches and report our results for both \textbf{D2-City} and \textbf{Visdrone} datasets.
For a fair comparison, the two baseline methods -- cloud-only and edge-only -- will also use the same cloud/edge object detection frequency.

Figure~\ref{fig:cloudpro}(a) compares  baseline algorithms with \pname{} (i.e. cloud-edge) with respect to the object detection accuracy (mAP@0.5) for D2City dataset. 
We create distinct pairs of object model combinations --- one running at the edge and the other on the cloud. 
Specifically, we evaluate using two edge models and three cloud models, a total of six combination pairs.
Our results show that \pname{} outperforms the edge-only and cloud-only baselines by 20-40\% for all combination pairs. Different object detection models exhibit different kind of errors, due to their DNN architectural design decisions, and \pname{} is able to combine these detections to reduce overall error and improve performance. This is akin to using an ensemble of cascading detection models in tandem to reduce error. 
We also observe that our approach's mAP is marginally better than the scenario where edge models are executed on every frame (i.e., \emph{ef-edge-det}), where no latency constraints on edge device is assumed. 
Even when compared to this impractical scenario, we observe that having redundant computation helps to improve accuracy over just using edge models.
In particular, the cloud-edge pair of CenterNet and SSD MobileNetv2 achieves the best performance. 




Figure~\ref{fig:cloudpro}(b) shows the same comparison using the Visdrone dataset. As noted in prior studies, object detection in this dataset is challenging, and models tend to have low mAP values~\cite{zhu2018visdrone}. 
Our results show that \pname{} achieves higher accuracy and outperforms baseline approaches by 20-50\%. 
This indicates that the combination of low accuracy yet computationally cheap \black{models} and high accuracy models but computationally expensive \black{models} can help further boost 
\black{accuracy over using just the former}. 
We also observe that the pair of RetinaNet and TinyYolo outperforms all baseline techniques. 





\textbf{Key Observations: } \textit{\pname{} outperforms baseline algorithms by as much as 50\%.
The performance of lower accuracy models can be improved using our edge-cloud fusion algorithm by combining results from higher accuracy models.
}


\subsection{\pname{} Tradeoff Analysis}
\subsubsection{Impact of Detection Frequency}
\label{subsec:detectfreq}
Most resource-constrained systems cannot execute deep learning-based object detections on each frame. Typically, the object detector runs only once every few frames and a lightweight object tracking is performed on intermediate frames. 
However, the accuracy of object tracking algorithms is poor and performance tends to degrade over time, especially on longer video sequences. Since infrequent detections at the edge and cloud may degrade performance, we assess
its impact on the overall accuracy of the system.

For our evaluation, we set the detection frequency and invoke edge and cloud models every $X$ number of frames and use RetinaNet as our cloud model and SSD MobileNetv2 as our edge model.  Figure~\ref{fig:heat} shows a heatmap indicating the accuracy of \pname{} using different edge and cloud detection frequencies. As expected, running more detections improves accuracy as it mitigates the degradation effects of object tracking. Moreover, if the scene changes frequently, the cloud detections may be stale, which may further contribute to degraded performance. And thus, invoking frequent detections at the edge helps in mitigating these effects.   

We can also tradeoff \black{flexibly} computation at the edge with \black{that in the} cloud. 
In particular, we can reduce the frequency at the edge (or cloud) and increase at the cloud (or edge) with little impact on accuracy. For example, running edge detections every 5th frame and cloud detections every 100th frame results in mAP@0.5 of 22.7. However,  we can instead trade-off computation and reduce the detection frequency at the edge by a fourth (e.g., run every 20th frame) and slighly more than triple the cloud frequency (e.g., every 30th frame) to achieve a similar accuracy (mAP@0.5=22.8). Such a scenario is quite common in edge devices where excess heat generated by running detectors often might result in throttling. If cloud resources are at a premium, we can get similar accuracy (mAP@0.5=22.5) with the edge and the cloud frequencies set to \black{every} 15th and 60th frame, respectively. Such flexibility allows application developers to perform tradeoffs to optimize for specific objectives.
These changes to cloud and edge detection frequencies to maintain similar accuracy also highlight the resilience of \pname{} to network variability. Reducing cloud detections forced by lower bandwidth can be compensated with higher edge detections.   

\subsubsection{Diagnostic Error Analysis}
\label{subsec:error-analysis}

As noted earlier in Section~\ref{subsec:baseline-comp}, 
\pname{} outperforms the baseline algorithms and also improves on the upper bound performance of using edge detections on each frame. 
However, mAP alone does not explain the effect of the various system parameters and the tradeoffs they introduce. To this end, we use TIDE~\cite{bolya2020tide}, a toolbox that helps disambiguate between six error types in object detection (Cls: classification error; Loc: localization error; Both: both cls and loc error; Dupe: duplicate predictions error; Bkg: background error; Miss: missed detections error).




\begin{figure}[t]
    \centering
    
    \includegraphics[width=0.6\columnwidth]{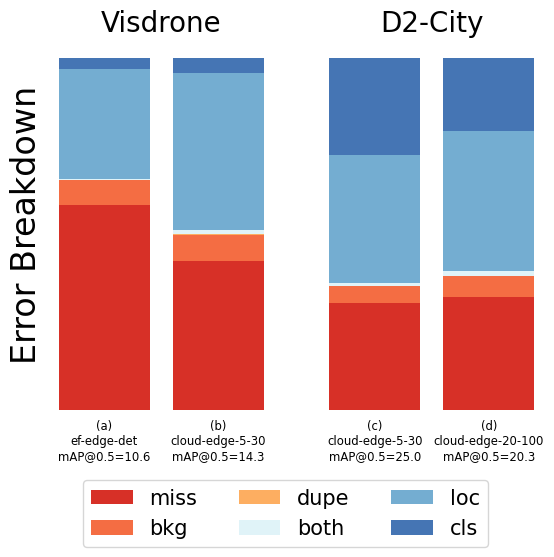}
    \caption{Error breakdowns on the two datasets for different \pname{} configurations}
    \label{fig:error_analysis_cloud_edge}
    \vspace{-0.2in}
\end{figure}
    
We analyze the error breakdown of \pname{} at different detection frequencies for the tinyYOLO-RetinaNet combination (like Section~\ref{subsec:baseline-comp}). It's clear from Figure~\ref{fig:error_analysis_cloud_edge} ((b) and (c)) that the kind of errors made by \pname{} on the two datasets are very different. On D2-City dataset, we see a substantially larger ratio of classification (class label mismatch) errors compared to VisDrone dataset, and a smaller ratio of missed detections. Thus, target domain is an important aspect in discussion of system tradeoffs.

Developers can adjust \pname{}'s parameters, such as changing cloud/edge detection frequency to reduce localization errors or missing detections. Depending on the scenario, one kind of error can be considered costlier than another (e.g., missing detection of ``person'' objects could be more problematic than mislabeling a ``van'' as a ``car''). On the Visdrone dataset, the ratio of missed detections is substantially lower \textit{(ef-edge-det (a) vs edge-cloud-5-30 (b))} contributing to increase in mAP from 10.6 mAP@0.5 to 14.3 mAP@0.5. This indicates is that the cloud models help in detecting objects that edge models are not able to detect. Next, on D2-City dataset, the ratio of localization errors \textit{(edge-cloud-5-30 (c) vs edge-cloud-20-100 (d))} increases as the overall mAP decreases from 25 mAP@0.5 to 20.3 mAP@0.5 with the decrease in cloud and edge detection frequency (from (5,30) to (20, 100)). However, if localization errors are tolerable in a use-case (e.g., counting scenarios), then savings in cloud cost and energy on the edge device can be made.



\textbf{Key Observations}: \textit{The flexibility to adjust detection frequency 
can immensely help resource-constrained scenarios. \pname{} provides the flexibility to tradeoff computation at the edge and cloud, while achieving similar performance. \pname{} can further mitigate different types of errors by changing system parameters and iterating on specific performance bottlenecks.} 


\begin{figure}[t]
    \begin{tabular}{cc}
    \includegraphics[width=1.4in]{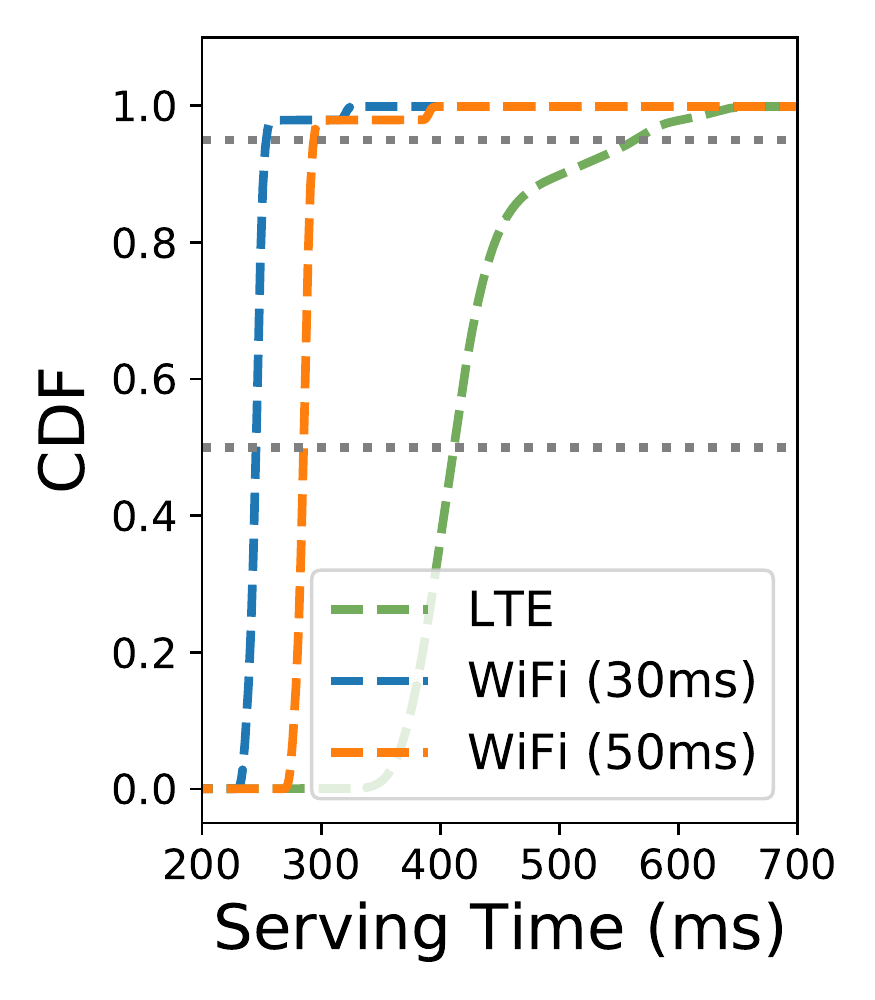} &
    \includegraphics[width=1.4in]{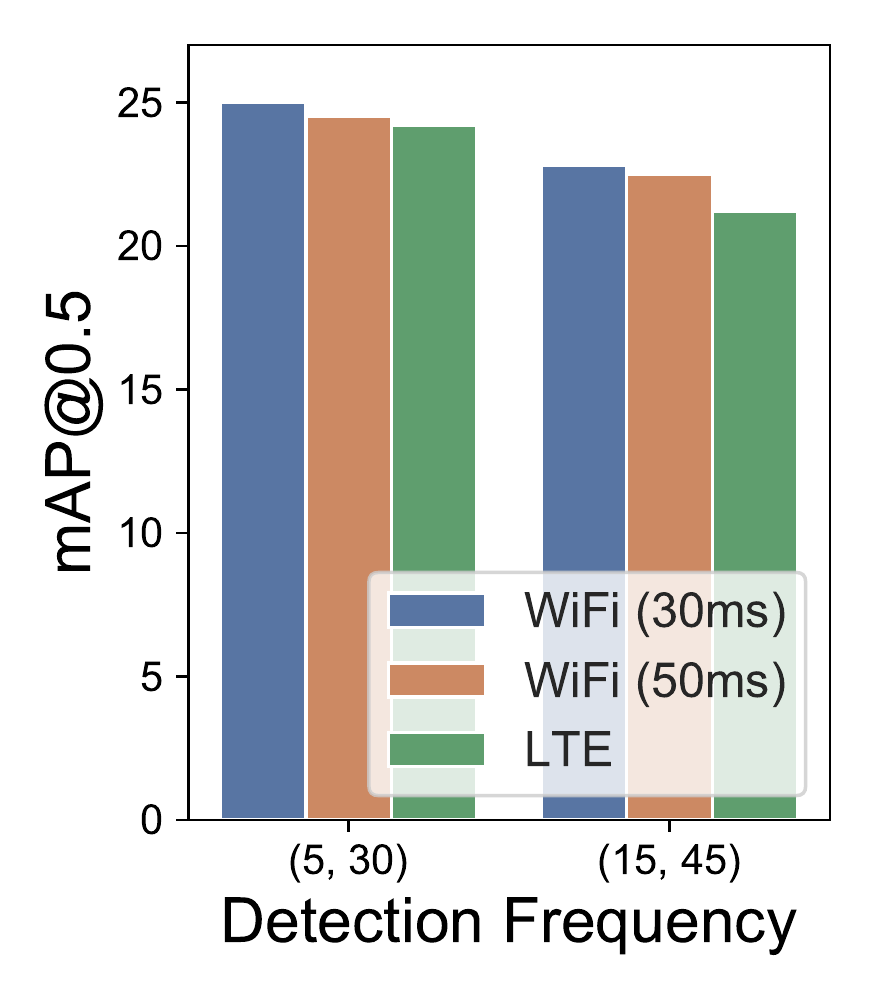}\\
    (a) CDF  & (b) Accuracy \\
    \end{tabular}
    \caption{Impact of network latency on accuracy for different network conditions.}
    \label{fig:network-exp}
    \vspace{-0.29in}
    \end{figure}

\begin{figure*}[t]
\minipage{0.32\textwidth}
    \centering
     \includegraphics[width=2in]{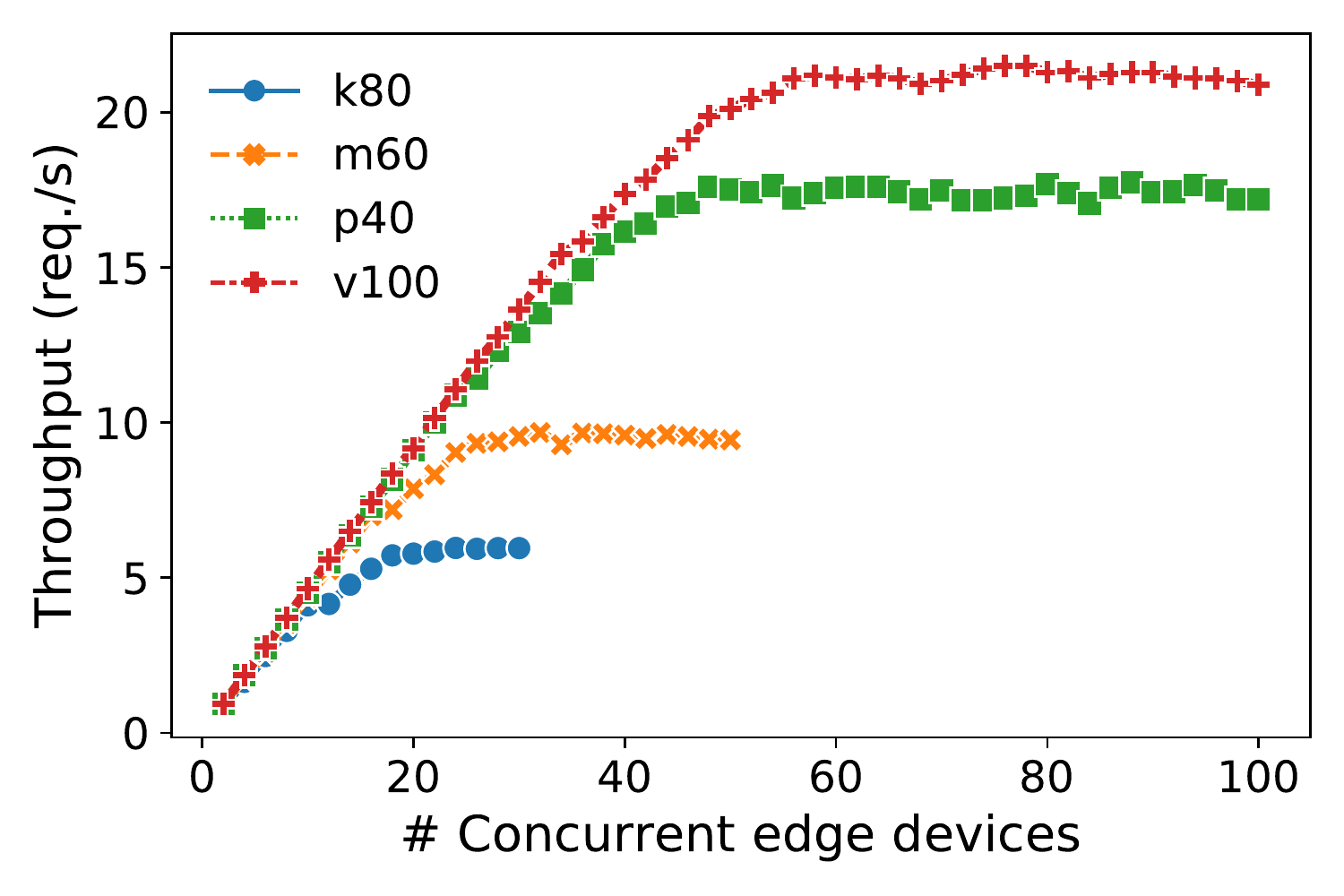}
    \caption{Throughput vs \#concurrent edge devices for different GPUs.}
    \label{fig:throughput}
\endminipage\hfill
    \minipage{0.32\textwidth}
    \centering
    \includegraphics[width=2in]{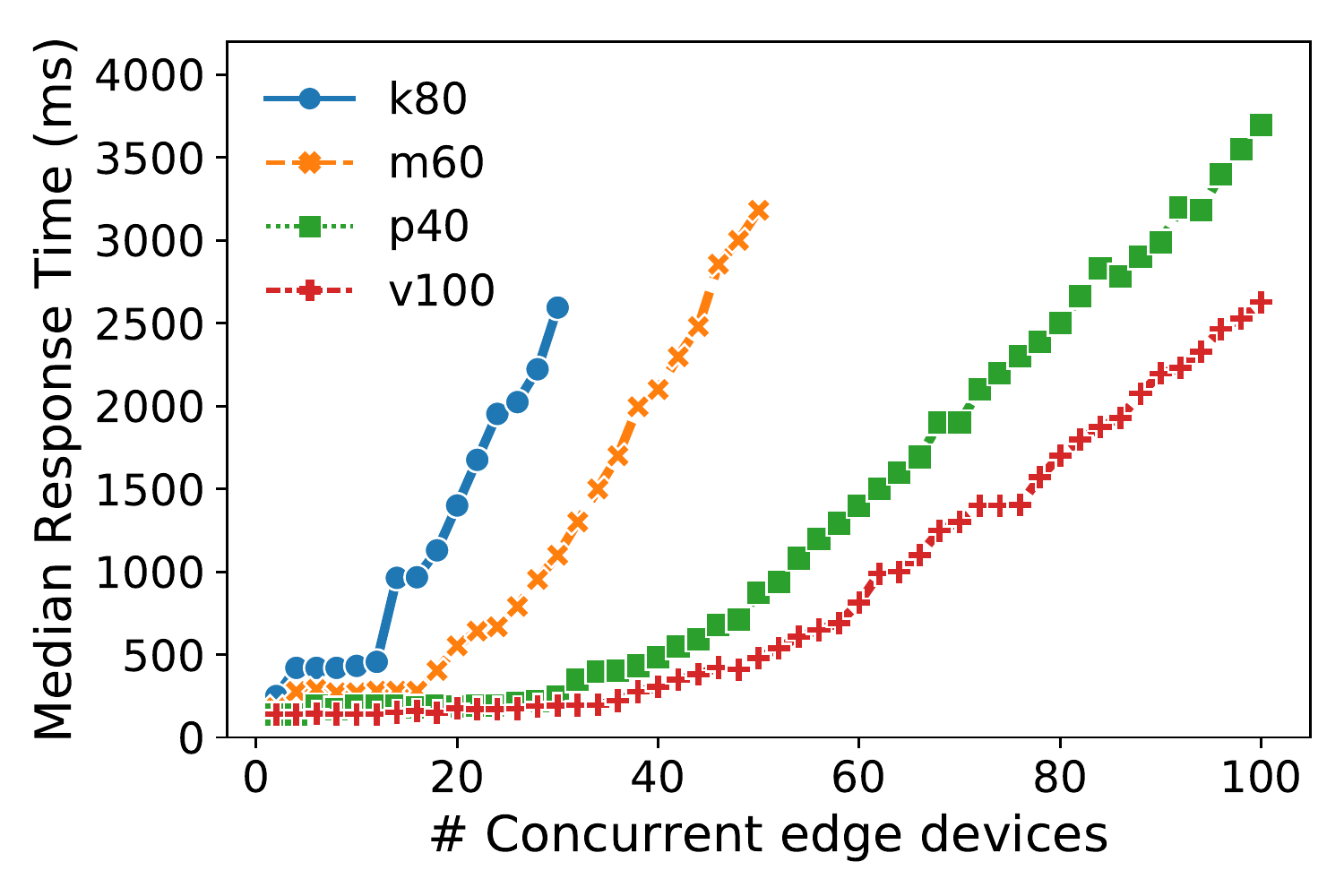}
     \caption{$50^{th}$ percentile response time vs \# concurrent edge devices}
     \label{fig:median}
    \endminipage\hfill 
\minipage{0.32\textwidth}
    \centering
    \includegraphics[width=2in]{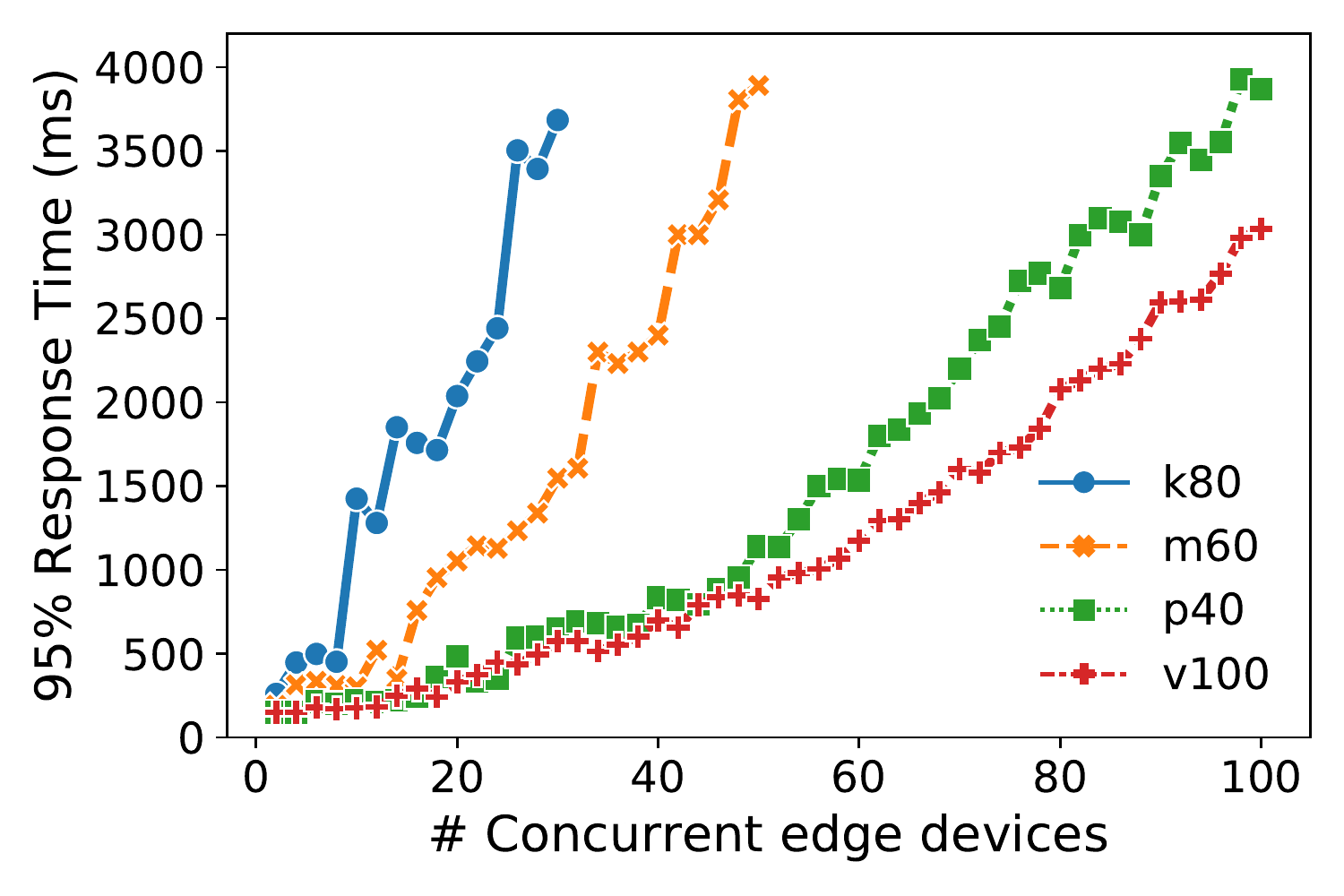}
    \caption{$95^{th}$ percentile response time vs \#concurrent edge devices}
    \label{fig:95_timescale}
\endminipage
\end{figure*}

\subsection{Impact of Network }

As discussed earlier, \pname{} receives responses asynchronously from the cloud and merges its annotations with the edge detections. Clearly, stale cloud responses affect accuracy. There are three factors that affect the \textit{serving time} of responses from the cloud --- (i) time to transmit a frame from the edge to the cloud, (ii) time to run inference on the frame at the cloud, and (iii) time to send the annotations from the cloud to the edge devices. Thus, we experiment with different networks to gauge their impact on the overall accuracy. We restrict our evaluation to the D2-City dataset with tinyYOLO and RetinaNet models running at the edge and the cloud, respectively. 

Figure~\ref{fig:network-exp}(a) show the cumulative distribution function (CDF) of the serving times observed on the four network conditions. The two gray-colored horizontal lines represent $50^{th}$ and the $95^{th}$ percentiles. Using WiFi (30ms), we get the lowest serving time, i.e., around 260 ms ($95^{th}$ percentile). Whereas, LTE has a significantly longer serving time compared to others (420 ms for $50^{th}$ and 570 ms for $95^{th}$ percentile). Unlike other network types, LTE also has a much higher standard deviation. The Figure~\ref{fig:network-exp}(b) shows the accuracy associated with the use of the four network types. Intuitively, accuracy degrades as serving times increase. This is because a change in the scene may render the stale output from the cloud useless. Thus, in the worst case, dynamic scenes where objects change frequently, such scenarios may not be able to take advantage of cloud resources.  
We observe this pattern in our analysis, where higher delays in serving time reduce accuracy. 
In particular, the model accuracy with LTE is the lowest at 21.1 --- i.e., 7\% lower than WiFi (30ms) in case of edge-cloud detection frequency at (15,45). 


\textbf{Key Observations: } 
\textit{\pname{} performance is sensitive to different network conditions. Specifically, a 310 ms difference in $95^{th}$ percentile serving time in network type results in 7\% reduction in accuracy.}


\subsection{\pname{}'s Scalability}


Next, we discuss how the added cost of additional cloud resources be amortized over many edge devices sharing the same \pname{} model server. 

To evaluate the scalability of \pname{} Model Server, we looked at four different generations of GPUs (i.e., K80, M60, P40, and V100) available on the cloud platforms. 
Consequently, we selected Microsoft Azure Ubuntu 18.04 VMs NC6v1 (K80), NV6v3 (M60), ND6v1 (P40), and NC6v3 (V100). 
As the two datasets consisted of various image resolutions, we choose a consistent image size ($512\times512$) for a fair comparison. We set the inference max batch size to 4 and use the Faster-RCNN model for the results discussed here (our most expensive cloud model). We benchmark using the HTTPS/JSON endpoint and define the payload and user characteristics using the \textit{Locust load testing library}~\cite{locust_library}. We looked at a scenario where the edge devices send requests once every 2 seconds (once every 60 frames). The payload involved adding users at a uniformly random rate of 3 edge devices per second until we reached the maximum desired number. Specifically, we varied  the concurrent number of edge devices sending requests between 2 to 100. 

Figure~\ref{fig:throughput} shows the throughput of the serving platform with a varying number of edge devices for the different GPU VMs. For a smaller number of devices, the GPUs are underutilized, and the throughput increases. However,  each of the four GPUs will hit a maximum throughput level with the increasing number of edge devices.  For newer GPU devices, such as V100 and P40, we get a maximum throughput of over 17 requests per second (req./s). Throughput can be increased by batching requests with a timeout queue at the expense of average latency. Whereas, the performance of the K80 is the worst, with throughput maxing out at slightly over $5$ req./s. Thus, during lower traffic conditions, one can go with older GPUs available at a discount compared to newer ones (the pricing market is dynamic and demand based). 
However, the newer GPUs can provide $>3\times$ the performance.

If an application can tolerate a median latency of 500 ms for inference on the cloud, we can support up to 60+ concurrent devices at a time using the V100 GPU (see Figure~\ref{fig:median}). If we consider a Reserved VM with a V100 GPU\footnote{Cost of a 3 year reserved Azure VM is 0.979\$ an hour. See \url{https://azure.microsoft.com/en-us/pricing/details/virtual-machines/linux/}}, 
the cost is 1.63\cent/hr. per concurrent device. This is a conservative analysis due to our model choice --- detectors less expensive than Faster RCNN (like RetinaNet) can support greater number of concurrent devices. This number reduces to 44, 19, and 12 for P40, M60, and K80, respectively. 
For $95^{th}$ percentile case, V100 can support 33 concurrent devices (see Figure~\ref{fig:95_timescale}). 
Moreover, for many video analytics applications not all edge devices are operational at all times. For example, one might use an AR/MR app on a mobile device for just 20 minutes a day. Similarly, a dashcam-based driver-assist application will only be operated while driving (around one hour a day). The overall number of edge devices supported will be orders of magnitude greater than the concurrent devices supported. 


\textbf{Key Observations: } \textit{A single instance of the \pname{} Model Server can handle an excess of 60 concurrent edge devices. 
We can divide the cost overhead of the VMs across hundreds of edge devices as only a few devices are operated at any given time for several real-time video analytics applications.}

 

\subsection{Performance on Edge Devices}
We evaluate the feasibility of \pname{} on the Nvidia Jetson Xavier device with installed JetPack SDK. Specifically, we deploy \pname{} on the device and calculated the maximum FPS obtained for TinyYOLO edge model and the CSRT tracker employed in the \pname{} Edge Manager. We achieve an average detection rate of 26.1 fps for a video stream for an image resolution of $540\times360$. For streaming applications (30 fps), we cannot invoke detection very often. Additionally, our tracker algorithm achieved 36.66 fps (> 30fps). Thus, it is feasible to use \pname{} for many video analytics applications where object detection is a crucial block. 

\textbf{Key Observations: } \textit{It is feasible to run \pname{} on edge-class devices. Reducing object detection frequency at the edge (while increasing the cloud detection frequency) can offer opportunities to execute downstream tasks in the analytics pipeline.}

\section{Qualitative Results}

\begin{figure*}[h]
    \centering
    \includegraphics[width=1\linewidth]{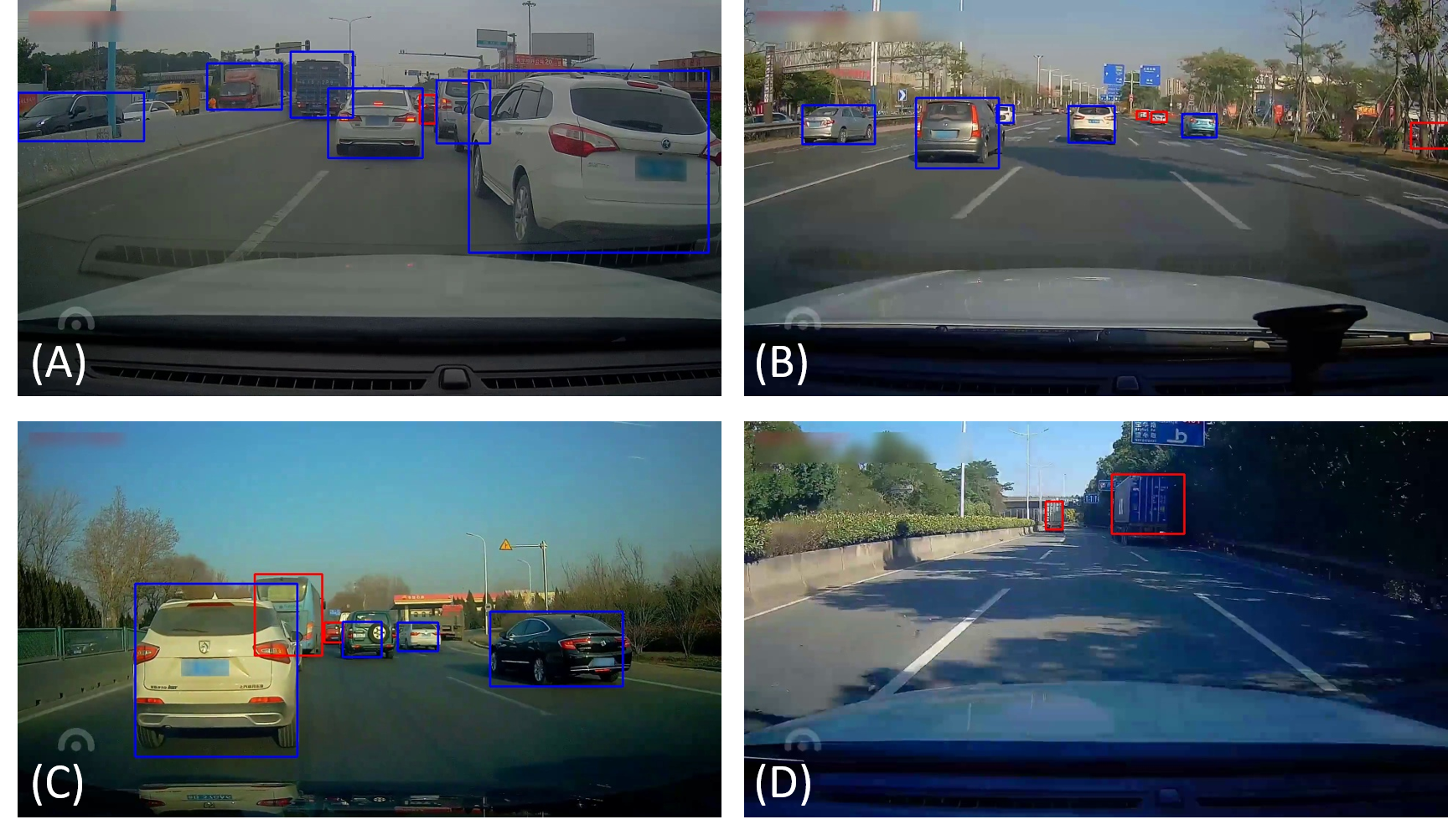}
    \caption{Detections on the D2City dataset. Cloud detections are colored red, whereas Edge detections are colored blue. The examples are representative.}
    \label{fig:visual_results_d2city}
\end{figure*}

\begin{figure*}[h]
    \centering
    \includegraphics[width=1\linewidth]{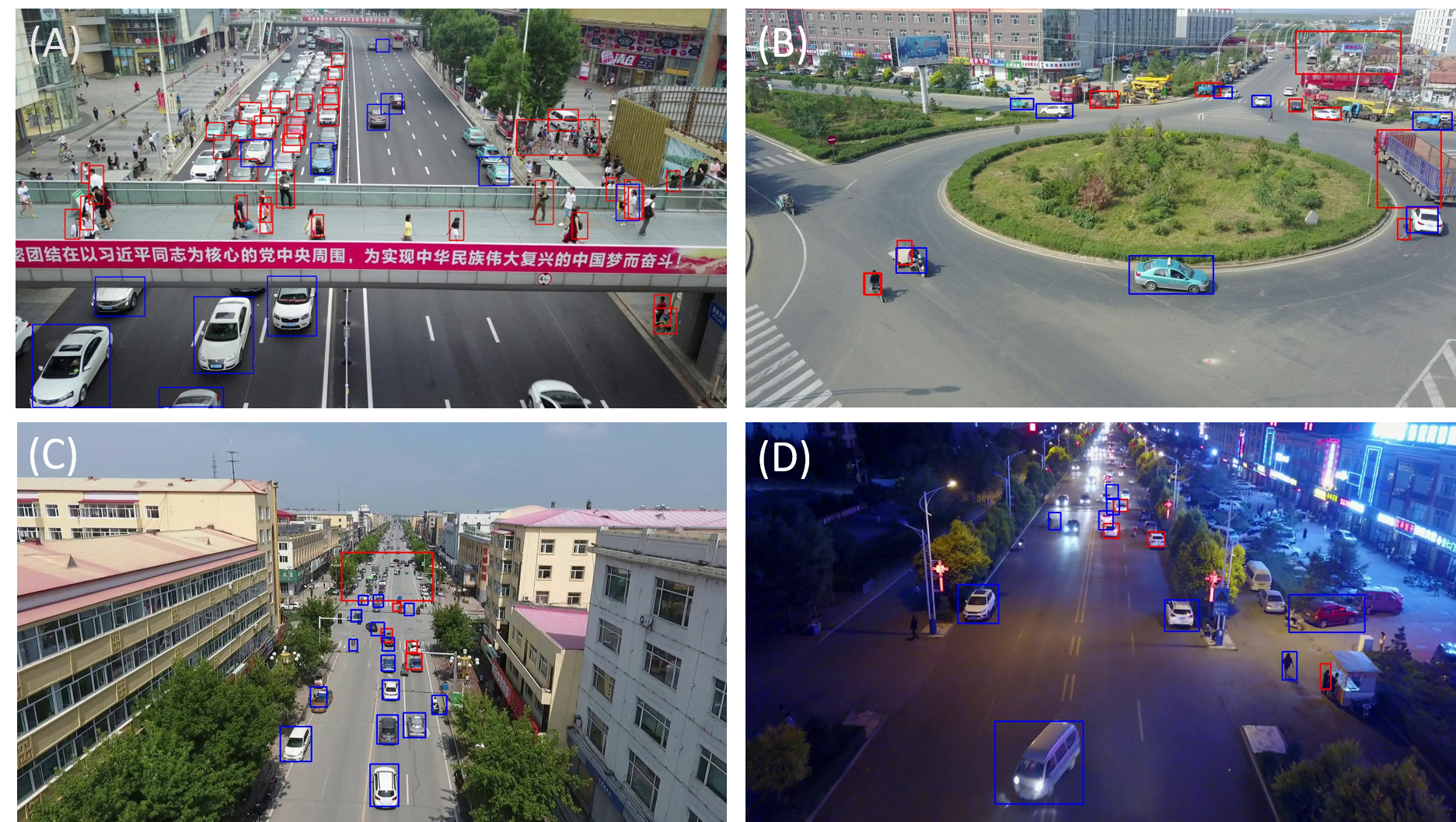}
    \caption{Detections on the Visdrone dataset. Cloud detections are colored red, whereas Edge detections are colored blue. The examples are representative.}
    \label{fig:visual_results_visdrone}
\end{figure*}

We visualize some frames from various sequences in the D2City dataset using \pname{} with TinyYOLO at the edge and RetinaNet at the cloud with a detection frequency of 5 and 30, respectively. As we can see, \pname{}'s Edge Cloud Fusion Algorithm helps in multiple scenarios. In Fig~\ref{fig:visual_results_d2city} (A), the edge model is able to identify and localize most of the objects, however, cloud model identifies a highly occluded car. While in Fig~\ref{fig:visual_results_d2city} (B), the cloud model is able to identify small objects (such as the \textit{cars} far away) which the edge model could not. The cloud model is able to identify the occluded \textit{bus}, which is close to the camera in Fig~\ref{fig:visual_results_d2city} (C). The edge model performs especially poorly in Fig~\ref{fig:visual_results_d2city} (D), as it's not able to identify any of the \textit{trucks} due to bad lighting conditions, which our cloud model can identify and localize correctly. This is consistent with prior observations that larger models are better at detecting small and occluded objects.Similar patterns emerge in VisDrone dataset, as observed in Fig~\ref{fig:visual_results_visdrone}. Moreover, as we can see in all the sub figures (specially in Fig~\ref{fig:visual_results_visdrone} (A)), the miss rate is significantly reduced by the detection of smaller objects by the cloud model. This is consistent with our observations in Section~\ref{subsec:error-analysis}.

\section{Related Work}



In this section, we contextualize our work with other studies.   

\noindent

\noindent
{\bf ML Model Optimizations: } There have been a few major ways of optimizing models themselves to reduce the inference time on the resource-constrained edge devices --- model pruning~\cite{han2015deep}, quantization~\cite{hubara2017quantized}, distillation~\cite{bajestani2020tkd} and hardware-aware neural architecture search~\cite{zhang2020skynet}. Unfortunately, the improvements in latency largely come at a cost of lower accuracies and generalization. Our approach is complementary to these approaches 
as we expect the performance arbitrage to exist and our results show that fusing the output can improve the overall accuracy. Moreover, any complementary improvement in the performance of small models reduces the dependence on the cloud for inference, increasing the concurrent clients our system can support.

\noindent
{\bf Video Analytics Optimizations: }
Live video analytics is emerging as an increasingly important problem because of its applications in multiple domains~\cite{ananthanarayanan2019demo}. 
However, providing efficient video inference remains a challenge due to constraints in compute, latency and bandwidth. As such, several studies have looked at optimizing several aspects within the video analytics pipeline to improve  overall performance~\cite{chen2015glimpse,apicharttrisorn2019frugal}.
Several papers have considered offloading the analysis to the cloud~\cite{ashok2015enabling,jiang2018chameleon}. 
Studies that offloads work to the cloud assumes that there is no stringent latency requirements. Some of them focus on optimizing video queries by selecting appropriate neural network and video configurations to save compute resources~\cite{jiang2018chameleon}. 
Separately, there have been several recent efforts to partition models across the cloud and edge~\cite{kang2017neurosurgeon}. Such techniques are not suitable for live analytics because the final result is primarily computed in the cloud, which increases the overall latency. Reducto~\cite{li2020reducto} investigates on-camera filtering, and dynamically adapts filtering decisions according to the time-varying
correlations. This is complementary to our work and can be used to reduce our edge detection frequency further.

\noindent
{\bf Object Detection Optimizations: }
There has also been studies that looks at leveraging both on-board compute and/or cloud resources to improve object detection~\cite{liu2019edge, apicharttrisorn2019frugal,chen2015glimpse}.  RedEye~\cite{likamwa2016redeye} performs early CNN computation in the analog domain on the image sensor. Apicharttrisorn et. al. ~\cite{apicharttrisorn2019frugal} proposed a detection technique for mobile-based AR applications that switches between lightweight object tracking and DNNs for object detection. 
Chen et al.~\cite{chen2015glimpse} presents a real-time object recognition pipeline that does object tracking locally but offloads DNN-based object detection to the cloud. DeepDecision~\cite{ran2018deepdecision} is measurement driven framework that considers running an object detector on the cloud or the edge depending on network conditions and edge hardware constraints. In contrast to prior, we perform redundant DNN-based detection both at the cloud and edge. Our analysis shows that such redundant inference from small DNN at the edge and large DNN at the cloud can significantly improve accuracy compared to baseline techniques that are based on existing work.

\section{Discussion and Future Work}


\noindent 
\textbf{Flexibility: } While we evaluate network latency and analyze the impact of detection frequency on edge, network bandwidth is also important. Since our approach allows the flexibility to change cloud detection frequency, we can control the data sent across the network to conserve bandwidth.  However, we can still achieve similar accuracy by increasing the detection frequency at the edge. Thus, users of \texttt{REACT} can achieve comparable accuracy by choosing a wide range of system parameters while satisfying use-case specific constraints, such as limited bandwidth or edge GPU cycles. Further, the modular design of \texttt{REACT} allows developers to swap models at the edge or the cloud as and when newer and improved DNN architectures are available. Our system also allows developers to choose a model serving system of their choice. 

\noindent 
\textbf{Generalizability:} Even though we evaluate our system on object detection tasks, we expect our approach to also work on human pose-estimation or instance segmentation applications. 
For example, human pose-estimation applications require instantaneous feedback for sports and dance activities
and to understand full-body sign language
--- all of which requires low latency analysis. Similarly, there is a need for low latency in instance segmentation tasks, such as for security and surveillance applications using robots.
Currently, these approaches need cloud-based resources. Our future work will involve extending our system to work for such applications. 

\noindent \textbf{Adaptive parameter setting:} We note that the detection frequency was fixed for our evaluation to show trade-off opportunities. However, the detection frequency can be adaptive and change based on variations in scene dynamism. For example, if the scene changes less frequently, we can decrease the detection frequency at the edge and/or the cloud to keep up with the desired accuracy. 
Detection frequency can also change due to systems constraints. If there is limited cloud resource available, one can reduce the cloud detection frequency. When cloud resources are cheap, increasing the cloud detection frequency can improve detection accuracy. Likewise, if the edge device experiences thermal throttling or is constrained by power consumption, then lowering edge detection frequency is necessary (say for battery-operated drones). Concurrent work~\cite{ghosh2021adaptive} has shown the feasibility of learning configurations for live streaming applications via Reinforcement Learning.

\section{Conclusion}

In this work, we introduced a new approach for improving live video analytics applications to leverage inferences available asynchronously from the cloud. \texttt{REACT} utilizes higher accuracy object detections from cloud to improve the past edge detections and cascade these to current predictions on the edge. \texttt{REACT} is performant, flexible, resilient to network latency, cost-effective and scalable. 
Our approach can be utilized by application developers to boost the performance in a variety of edge video analytics applications.

\bibliographystyle{ACM-Reference-Format}
\bibliography{paper}


\end{document}